\documentclass[10pt,twocolumn,letterpaper]{article}

\usepackage{iccv}
\usepackage{times}
\usepackage{epsfig}
\usepackage{graphicx}
\usepackage{amsmath}
\usepackage{amssymb}
\usepackage{booktabs}
\usepackage{marvosym}
\usepackage{url}

\usepackage[pagebackref=true,breaklinks=true,colorlinks,bookmarks=false]{hyperref}

\usepackage[capitalize]{cleveref}
\crefname{section}{Sec.}{Secs.}
\Crefname{section}{Section}{Sections}
\Crefname{table}{Table}{Tables}
\crefname{table}{Tab.}{Tabs.}

\usepackage{multirow}
\usepackage{colortbl}
\usepackage{xcolor}

\usepackage{pifont}
\newcommand{\cmark}{\ding{51}}%
\newcommand{\xmark}{\ding{55}}%

\usepackage[accsupp]{axessibility}  % Improves PDF readability for those with disabilities.

\iccvfinalcopy

\ificcvfinal\fi

\begin{document}

%%%%%%%%% TITLE
\title{Adaptive Rotated Convolution for Rotated Object Detection}

\author{
Yifan Pu$^1$\thanks{Equal Contribution. \qquad \Letter \  Corresponding author.}\qquad 
Yiru Wang$^2$\footnotemark[1]\qquad
Zhuofan Xia$^1$\qquad
Yizeng Han$^1$\qquad
Yulin Wang$^1$\\
Weihao Gan$^3$\qquad
Zidong Wang$^1$\qquad
Shiji Song$^1$\qquad
Gao Huang$^{1,4}$\textsuperscript{\Letter} \\ % \thanks{Corresponding Author.}
\normalsize{
$^1$Department of Automation, BNRist, Tsinghua University} \qquad
\normalsize{
$^2$SenseTime Research} \\
\normalsize{
$^3$Mashang Consumer Finance Co., Ltd.
}\;
\normalsize{
$^4$Beijing Academy of Artificial Intelligence
} \\
{\tt\small pyf20@mails.tsinghua.edu.cn, wangyiru@sensetime.com,} 
{\tt\small gaohuang@tsinghua.edu.cn}
}

\maketitle
\ificcvfinal\thispagestyle{empty}\fi

%%%%%%%%% ABSTRACT
\begin{abstract}

Rotated object detection aims to identify and locate objects in images with arbitrary orientation. In this scenario, the oriented directions of objects vary considerably across different images, while multiple orientations of objects exist within an image. This intrinsic characteristic makes it challenging for standard backbone networks to extract high-quality features of these arbitrarily orientated objects.
In this paper, we present Adaptive Rotated Convolution (ARC) module to handle the aforementioned challenges. 
In our ARC module, the convolution kernels rotate adaptively to extract object features with varying orientations in different images, and an efficient conditional computation mechanism is introduced to accommodate the large orientation variations of objects within an image.
The two designs work seamlessly in rotated object detection problem. Moreover, ARC can conveniently serve as a plug-and-play module in various vision backbones to boost their representation ability to detect oriented objects accurately.
Experiments on commonly used benchmarks (DOTA and HRSC2016) demonstrate that equipped with our proposed ARC module in the backbone network, the performance of multiple popular oriented object detectors is
significantly improved (\eg +3.03\% mAP on Rotated RetinaNet and +4.16\% on CFA).
Combined with the highly competitive method Oriented R-CNN, the proposed approach achieves state-of-the-art performance on the DOTA dataset with 81.77\% mAP.
Code is available at \url{https://github.com/LeapLabTHU/ARC}.
\vspace{-0.5cm}

\end{abstract}

%%%%%%%%% BODY TEXT
\section{Introduction}
\label{sec:intro}

% Establishing a territory

Rotated object detection has become an emerging research topic in recent years \cite{xia2018dota, yang2019scrdet, qian2021learning}. Different from generic object detection, where object instances are assumed to be aligned with the image axes, objects in the natural scenes are prone to be placed with arbitrary orientation.
This phenomenon is commonly observed in the field of scene text detection~\cite{yao2012detecting, karatzas2013icdar, karatzas2015icdar}, face detection~\cite{jain2010fddb, yang2016wider}, and aerial image recognition\cite{heitz2008learning, liu2016ship, xia2018dota}, \etc.
In particular, in {Embodied AI}~\cite{duan2022survey, xie2023learning, lv2023learning, yan2022adaptive, franklin1997autonomous} tasks, as the agents need to explore and interact with the environment~\cite{bohg2017interactive, bajcsy2018revisiting, lee2022uncertainty}, the captured images may contain an object from arbitrary viewpoints~\cite{fujita2020distributed, sock2020active}. This poses great challenges for detection algorithms to accurately locate oriented objects.

\begin{figure}[t]
  \centering
  \includegraphics[width=0.99\linewidth]{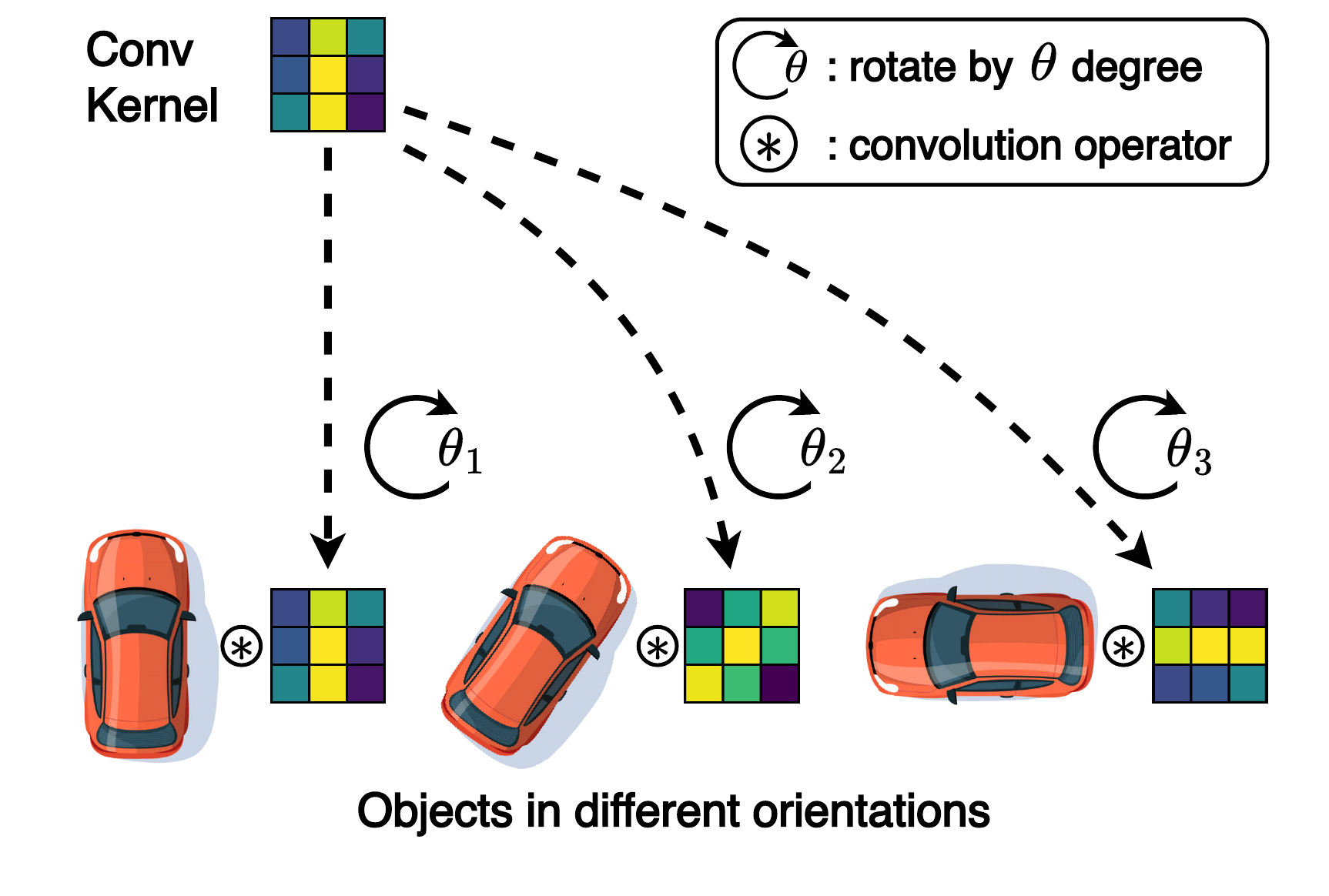}

  \caption{{\bf The motivation of our work}. In the rotated object detection scenario, object instances with similar visual appearance are placed with arbitrary orientation (\eg, the cars). As a result, it is reasonable to rotate the convolution kernels according to the orientation of the objects in a data-dependent manner rather than processing the image samples with the same static kernel.}
  \label{fig:motivation}
  \vspace{-0.5cm}
\end{figure}

%% current works
Recently, considerable progress has been achieved  in detecting rotated objects. For instance, various rotated object representations \cite{xu2020gliding, guo2021beyond, li2022oriented, hou2022g, yang2022detecting} and more suitable loss functions for these object representations~\cite{qian2021learning, chen2020piou, yang2021rethinking, yang2021learning, yang2022kfiou} have been extensively analyzed. The mechanisms of the rotated region of interest extraction~\cite{ding2019learning, xie2021oriented} and label assignment strategies~\cite{ming2021sparse, ming2021dynamic} have also been well explored. Moreover, the structure of the detection networks, including the {neck}~\cite{yang2019scrdet, yang2021r3det, yang2022scrdet++} and {head}~\cite{han2021align, hou2022shape} of detectors, and rotated region proposal networks~\cite{xie2021oriented, cheng2022anchor} has been comprehensively studied as well. However, little effort has been made in the design of a proper \emph{backbone} feature extractor.

The quality of the features extracted by backbone networks is crucial to many vision tasks. In particular, rotated object detection raises great challenges for backbone network design, since the orientation of objects varies \emph{across different images}. In the meanwhile, multiple orientations of objects also exist \emph{within an image}. Despite the significant differences between the images with generic items and those with oriented objects, the design of conventional visual backbones has mostly ignored the inherent characteristics. Therefore, the architecture of standard backbone models may be sub-optimal in the rotated object detection task.

In this paper, we address the above challenges by proposing a simple yet effective \textbf{Adaptively Rotated Convolution (ARC)} module. In this module, the convolution kernel \emph{adaptively rotates} to adjust the parameter conditioned on each input (\cref{fig:motivation}), where the rotation angle is predicted by a routing function in a \emph{data-dependent} manner.
Furthermore, an efficient conditional computation technique is employed, which endows the detector with more adaptability to handle objects with various orientations within an image. Specifically, multiple kernels are rotated individually and then combined together before being applied for convolution operations. This combine-and-compute procedure is equivalent to performing convolution with different kernels separately and then summing up the obtained results, yet the computation could be significantly reduced \cite{han2021dynamic}. The two designs work seamlessly, effectively enlarging the parameter space and elegantly endowing the network with more flexibility to detect objects in different orientations.

The proposed ARC module can conveniently serve as a plug-and-play module in convolution layers with arbitrary kernel size. As a result, any backbone network with convolution layers can enjoy the powerful representation ability of rotated objects by using the ARC module. For example, we can replace the convolution layers with the ARC module in the commonly used backbone network ResNet~\cite{he2016deep} to build the proposed backbone network ARC-ResNet.

We evaluate our method on two popular rotated object detection benchmarks (DOTA~\cite{xia2018dota} and HRSC2016~\cite{liu2016ship}). Extensive experiments validate that with the backbone network equipped with our proposed adaptive rotated convolution (ARC) module, the performance of various popular oriented object detectors can be effectively enhanced on both datasets. When combined with a highly competitive method Oriented R-CNN~\cite{xie2021oriented}, our approach achieves state-of-the-art performance on the DOTA-v1.0 benchmark.

\section{Related work}
\label{sec:related}

\noindent
{\bf Rotated object detection} attempts to extend generic horizontal detection to a finer-grained problem by introducing the oriented bounding boxes. Along this path, one line of work is devoted to establishing specialized rotated object detectors, including the feature refinement design in the detector neck~\cite{yang2019scrdet, yang2021r3det, yang2022scrdet++}, the oriented region proposal network~\cite{xie2021oriented, cheng2022anchor}, the rotated region of interest (RoI) extraction mechanism~\cite{ding2019learning, xie2021oriented}, detector head design~\cite{han2021align, hou2022shape} and advanced label assignment strategy~\cite{ming2021sparse, ming2021dynamic}. Another line of work focuses on designing more flexible object representations. For example, Oriented RepPoints~\cite{li2022oriented} represent objects as a set of sample points. Gliding Vertex~\cite{xu2020gliding} introduces a novel representation by adding four gliding offset variables to classical horizontal bounding box representation. CFA~\cite{guo2021beyond} models irregular object layout and shape as convex hulls. G-Rep~\cite{hou2022g} proposes a unified Gaussian representation to construct Gaussian distributions for oriented bounding box, quadrilateral bounding box, and point set. Meanwhile, a proper loss function for various oriented object representations has also been extensively studied. GWD~\cite{yang2021rethinking} and KLD~\cite{yang2021learning} convert the rotated bounding box into a 2-D Gaussian distribution, and then calculate the the Gaussian Wasserstein distance and Kullback-Leibler Divergence separately as losses. KFIoU~\cite{yang2022kfiou} proposes an effective approximate SkeIoU loss on Gaussian modeling and Kalman filter. Furthermore, some studies approach the problem from unique perspectives.~\cite{yang2020arbitrary}. 
% CSL~\cite{yang2020arbitrary} presents a new rotation prediction pipeline, which converts orientation prediction from a regression problem to a classification task, to avoid boundary discontinuity and square-like problems.

Albeit effective, the aforementioned works generally focus on the detector design, while the design of a proper backbone feature extractor is rarely explored. ReDet~\cite{han2021redet} incorporates rotation-equivariant operations~\cite{weiler2019general} into the backbone to produce rotation-equivariant features. Although the orientation information is reserved by rotation-equivariant operations, the variation of oriented objects within an image and across the dataset is ignored. In contrast, the proposed adaptive rotated convolution method embraces this characteristic of rotated objects.

\noindent
{\bf Dynamic network} is an emerging research topic in the deep learning community~\cite{han2021dynamic, wang2023computation}. In contrast to static network models, which share a fixed computation paradigm across different samples, dynamic networks can adapt their network structures or parameters to the input in the inference stage, and thus enjoy some favorable properties over static models such as efficiency, representation power, adaptiveness, compatibility, and interpretability. Dynamic networks can be divided into the following three categories: sample-wise, spatial-wise, and temporal-wise dynamic networks.

\begin{figure*}[ht]
  \centering
  \includegraphics[width=0.999\linewidth]{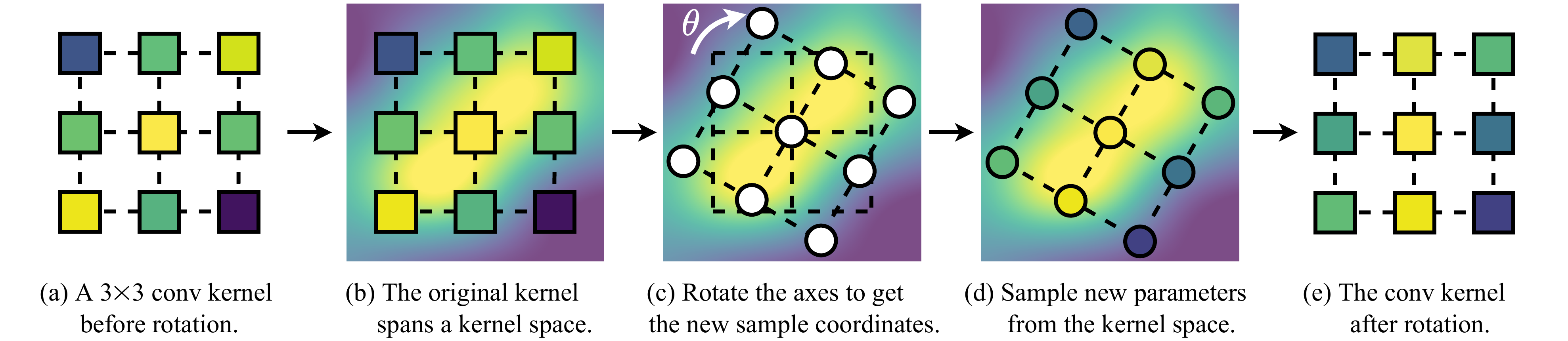}

  \caption{{\bf The procedure of rotating a $\bf{3\times3}$ convolution kernel.} (a) One channel of the original $3\times3$ convolution kernel. (b) By interpolation techniques, the 3$\times$3 weight values can span into a 2D kernel space. (c) Rotate the original coordinates to get the sample coordinates for the new rotated convolution kernel. (d) Sample weight values in the new rotated coordinates from the kernel space. (e) The rotated convolution kernel weights are obtained by sampling from the original space.}
  \label{fig:rotate}
  \vspace{-0.4cm}
\end{figure*}

Sample-wise dynamic networks process different inputs in a data-dependent manner and are typically designed from two perspectives: dynamic architecture and dynamic parameter. The former one generally adjusts model architecture to allocate appropriate computation based on each sample, therefore reducing redundant computation for increased efficiency. Popular techniques include early-exiting\cite{teerapittayanon2016branchynet, bolukbasi2017adaptive, huang2017multi, wang2021not, han2022learning, han2023dynamic}, layer skipping~\cite{wang2018skipnet, veit2018convolutional}, mixture of experts~\cite{shazeer2017outrageously, mullapudi2018hydranets, fedus2021switch}, and dynamic routing in supernets~\cite{tanno2019adaptive, li2020learning}. In contrast, dynamic parameter approach adapts network parameters to every input with fixed computational graphs, with the goal of boosting the representation power with minimal increase of computational cost. The parameter adaptation can be achieved
from four aspects: 1) adjusting the trained parameters based on the input~\cite{su2019pixel, yang2019condconv}; 2) directly generating the network parameters from the input~\cite{ha2016hypernetworks, jia2016dynamic, gao2019deformable, ma2020weightnet}; 3) adapting the kernel shape conditioned on the input~\cite{zhu2019deformable, dai2017deformable, xia2022vision}; and 4) rescaling the features with soft attention~\cite{wang2017residual, hu2018squeeze, he2023HQG}. In addition to the parameters during inference, training parameters can also be made dynamic~\cite{pu2023fine}. Spatial-wise dynamic networks perform spatially adaptive inference on the most informative regions, and reduce the unnecessary computation on less important areas. Existing works mainly include three levels of dynamic computation: resolution level \cite{yang2020resolution, zhu2021dynamic, He2023Camouflaged, he2023weaklysupervised, he2023strategic}, region level \cite{wang2020glance, huang2022glance} and pixel level \cite{dong2017more, verelst2020dynamic, xie2020spatially, han2022latency, han2023latency}. Temporal-wise dynamic networks extend the ideology of dynamic computation into the sequential data, typically on processing text data~\cite{hansen2019neural} and videos~\cite{wang2021adaptive}.

Our proposed method can be categorized into the parameter adjusting class. With adaptive rotated convolution kernel parameters, the proposed convolution module boosts the representation power of the backbone feature extractor, especially under the rotated object detection scenario.

\section{Method}
\label{sec:method}

In this section, we first introduce the convolution kernel rotation mechanism given a rotation angle $\theta$ (\cref{sec:met_rotate}). The network structure and the design methodology of the routing function are further presented (\cref{sec:met_router}). Finally, the overall picture of the proposed adaptive convolution module is illustrated in \cref{sec:met_module}. We also provide the implementation details of the ARC module in \cref{sec:met_imp}.

\subsection{Rotate the convolution kernels}
\label{sec:met_rotate}

% The standard convolution, which is adopted in the backbone network of most oriented object detectors, uses the same parameters to extract the feature of all image samples. That means in the rotated object detection scenario, the object instances with different rotation angles are processed with static convolution kernel heading into the same direction. To alleviate the gap between arbitrarily-oriented object instances and statically-oriented convolution kernels, we propose to rotate the convolution kernels by sampling weight in the kernel space in a data-dependent manner.
The standard convolution, adopted as the backbone of most oriented object detectors, employs consistent parameters to extract features from all image samples. In scenarios of rotated object detection, this implies that object instances, irrespective of their rotational angles, are processed using a static convolution kernel oriented in a fixed direction. To bridge the gap between arbitrarily-oriented object instances and these statically-oriented convolution kernels, we propose rotating the convolution kernels by sampling weights within the kernel space in a data-driven manner.

First, we illustrate how one channel of a convolution kernel rotates given a rotation angle $\theta$. We define the counter-clockwise direction as positive. Instead of treating the convolution weight as independent parameters (\cref{fig:rotate}{\color{red} (a)}), we treat them as sampled points from a kernel space. Therefore, the original convolution parameters can span a kernel space by interpolation (\cref{fig:rotate}{\color{red} (b)}). In practice, we use bilinear interpolation. The procedure of rotating a convolution kernel is the procedure of sampling new weight values in the rotated coordinates from the kernel space. The sample coordinates are obtained by revolving the original coordinates clockwise around the central point by $\theta$ degree (\cref{fig:rotate}{\color{red} (c)}). By sampling the values from the original kernel space in the rotated coordinates (\cref{fig:rotate}{\color{red} (d)}), the rotated convolution kernel is obtained (\cref{fig:rotate}{\color{red} (e)}). Note that to rotate the convolution kernel by $\theta$ degrees counter-clockwise, the coordinates in \cref{fig:rotate} need to take a $\theta$ degree clockwise rotation.

Now that the mechanism of rotating one channel of a convolution kernel (with a shape of $[k, k]$, $k$ is the kernel size) is given, the rotation procedure of the overall parameters of a convolution layer (with a shape of $[C_{\text out}, C_{\text in}, k, k]$, where $C_{\text in}$ and $C_{\text out}$ denote the number of input channels and the number of output channels, respectively) is easily extended. We simply apply the same procedure to all the $C_{\text in}$ channels and all the $C_{\text out}$ kernels to obtain the rotated weight parameter for a convolution layer.

\begin{figure*}[ht]
  \centering
  \includegraphics[width=\linewidth]{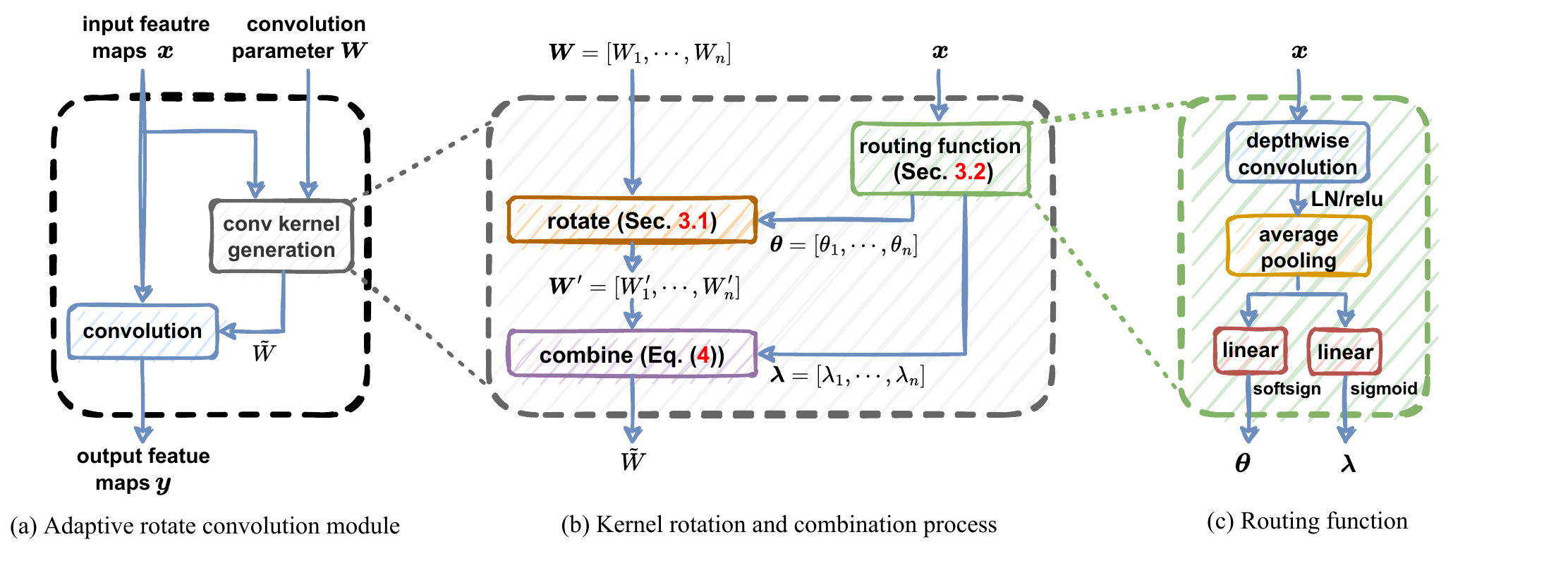}
  
  \caption{{\bf The illustration of the proposed adaptive rotated convolution module (ARC module).} (a) The macro view of the ARC module. The final convolution kernels $\tilde{W}$ are generated from the original convolution parameter $\boldsymbol{W}$, guided by the input feature map $\boldsymbol{x}$. (b) The process of convolution kernel generation. The kernels $\boldsymbol{W} = [W_1, \cdots, W_n]$ rotate $\boldsymbol{\theta} = [\theta_1, \cdots, \theta_n]$ degree and then combine together with the weight $\boldsymbol{\lambda} = [\lambda_1, \cdots, \lambda_n]$. The $\boldsymbol{\theta}$ and the $\boldsymbol{\lambda}$ are predicted by the routing function in a data-dependent manner. (c) The architecture of the routing function. The image feature $\boldsymbol{x}$ is encoded by a depthwise convolution with an average pooling layer followed behind. The $\boldsymbol{\theta}$ and the $\boldsymbol{\lambda}$ are predicted by two different branches with different activation functions, respectively.}
  \label{fig:module}
  \vspace{-0.3cm}
\end{figure*}

\subsection{Routing function}
\label{sec:met_router}

The routing function is one of the key components of the proposed adaptive rotated convolution module because it predicts the rotation angles and the combination weights in a data-dependent manner.
The routing function takes the image feature $\boldsymbol{x}$ as input and predicts a set of rotated angles $[\theta_1, \cdots, \theta_n]$ for the set of kernels and the corresponding combination weights $[\lambda_1, \cdots, \lambda_n]$ for them. 

The overall architecture of the routing function is shown in \cref{fig:module}{\color{red} (c)}. The input image feature $\boldsymbol{x}$, with a size of $[C_\text{in}, H, W]$, is first fed into a lightweight depthwise convolution with a 3$\times$3 kernel size, followed by a layer normalization~\cite{ba2016layer} and a ReLU activation. Then the activated feature is averaged pooled into a feature vector with $C_{\text in}$ dimensions. The pooled feature vector is passed into two different branches. The first branch is the rotation angle prediction branch, composed of a linear layer and a softsign activation. We set the bias of this linear layer as false to avoid learning biased angles. The softsign activation is adopted to have a low saturation speed. In addition, the output of softsign layer is multiplied by a coefficient to enlarge the range of rotation. The second branch, named combination weights prediction branch, is responsible for predicting the combination weights $ {\boldsymbol \lambda} $. It is constructed by a linear layer with bias and a sigmoid activation. The routing function is initialized from a zero-mean truncated normal distribution with 0.2 standard deviation to let the module produce small values at the start of the learning procedure.

% The proposed routing function is built on the shoulders of the design in Yang \etal~\cite{yang2019condconv}, except for two key components: (1) Spatial information encoding, which leverages an early lightweight depthwise convolution to extract the orientation information of image with arbitrarily oriented objects; (2) The rotation angle prediction branch, which is used to predict the angle for the experts to rotate.

\begin{table*}[ht]
\begin{center}
\resizebox{1.00\textwidth}{!}{
\setlength{\tabcolsep}{1.1mm}
\begin{tabular}{l|c|c c c c c c c c c c c c c c c|l}
    \toprule
    Method & Backbone & PL & BD & BR & GTF & SV & LV & SH & TC & BC & ST & SBF & RA & HA & SP & HC & $\ $ mAP \\
    
    \midrule
    \multicolumn{18}{l}{{\emph{Single-stage methods}}} \\
    \midrule
    Rotated & R50 & 89.29 & 78.54 & 41.13 & 66.29 & 76.92 & 61.68 & 77.40 & 90.89 & 81.37 & 82.89 & 59.05 & 63.79 & 55.05 & 61.95 & 40.14 & $\ $ 68.42 \\
    RetinaNet~\cite{lin2017focal} & \cellcolor{lightgray!30}{\bf ARC-R50} & \cellcolor{lightgray!30}89.59 & \cellcolor{lightgray!30}82.45 & \cellcolor{lightgray!30}41.66 & \cellcolor{lightgray!30}71.30 & \cellcolor{lightgray!30}77.71 & \cellcolor{lightgray!30}63.15 & \cellcolor{lightgray!30}78.04 & \cellcolor{lightgray!30}90.90 & \cellcolor{lightgray!30}84.95 & \cellcolor{lightgray!30}83.55 & \cellcolor{lightgray!30}57.74 & \cellcolor{lightgray!30}69.06 & \cellcolor{lightgray!30}54.72 & \cellcolor{lightgray!30}72.34 & \cellcolor{lightgray!30}54.58 & $ \  \; \cellcolor{lightgray!30}\textbf{71.45}_{(\textcolor{blue}{\uparrow \textbf{3.03}})}$ \\
    \cmidrule{1-18}
    
    \multirow{2}{*}{R3Det~\cite{yang2021r3det}}
	& R50 & 89.00 & 75.60 & 46.64 & 67.09 & 76.18 & 73.40 & 79.02 & 90.88 & 78.62 & 84.88 & 59.00 & 61.16 & 63.65 & 62.39 & 37.94 & $\ $ 69.70 \\
    & \cellcolor{lightgray!30}{\bf ARC-R50} & \cellcolor{lightgray!30}89.49 & \cellcolor{lightgray!30}78.04 & \cellcolor{lightgray!30}46.36 & \cellcolor{lightgray!30}68.89 & \cellcolor{lightgray!30}77.45 & \cellcolor{lightgray!30}72.87 & \cellcolor{lightgray!30}82.76 & \cellcolor{lightgray!30}90.90 & \cellcolor{lightgray!30}83.07 & \cellcolor{lightgray!30}84.89 & \cellcolor{lightgray!30}58.72 & \cellcolor{lightgray!30}68.61 & \cellcolor{lightgray!30}64.75 & \cellcolor{lightgray!30}68.39 & \cellcolor{lightgray!30}49.67 & $ \  \; \cellcolor{lightgray!30}\textbf{72.32}_{(\textcolor{blue}{\uparrow \textbf{2.62}})}$ \\
    \cmidrule{1-18}
    
    \multirow{2}{*}{S$^{2}$ANet~\cite{han2021align}}
	& R50 & 89.30 & 80.11 & 50.97 & 73.91 & 78.59 & 77.34 & 86.38 & 90.91 & 85.14 & 84.84 & 60.45 & 66.94 & 66.78 & 68.55 & 51.65 & $\ $ 74.13 \\
    & \cellcolor{lightgray!30}{\bf ARC-R50} & \cellcolor{lightgray!30}89.28 & \cellcolor{lightgray!30}78.77 & \cellcolor{lightgray!30}53.00 & \cellcolor{lightgray!30}72.44 & \cellcolor{lightgray!30}79.81 & \cellcolor{lightgray!30}77.84 & \cellcolor{lightgray!30}86.81 & \cellcolor{lightgray!30}90.88 & \cellcolor{lightgray!30}84.27 & \cellcolor{lightgray!30}86.20 & \cellcolor{lightgray!30}60.74 & \cellcolor{lightgray!30}68.97 & \cellcolor{lightgray!30}66.35 & \cellcolor{lightgray!30}71.25 & \cellcolor{lightgray!30}65.77 & $ \  \; \cellcolor{lightgray!30}\textbf{75.49}_{(\textcolor{blue}{\uparrow \textbf{1.36}})}$ \\
    
    \midrule
    \multicolumn{18}{l}{{\emph{Two-stage methods}}} \\
    \midrule
    Rotated Faster & R50 & 89.40 & 81.81 & 47.28 & 67.44 & 73.96 & 73.12 & 85.03 & 90.90 & 85.15 & 84.90 & 56.60 & 64.77 & 64.70 & 70.28 & 62.22 & $\ $ 73.17 \\
    R-CNN~\cite{ren2015faster} & \cellcolor{lightgray!30}{\bf ARC-R50} & \cellcolor{lightgray!30}89.49 & \cellcolor{lightgray!30}82.11 & \cellcolor{lightgray!30}51.02 & \cellcolor{lightgray!30}70.38 & \cellcolor{lightgray!30}79.07 & \cellcolor{lightgray!30}75.06 & \cellcolor{lightgray!30}86.18 & \cellcolor{lightgray!30}90.91 & \cellcolor{lightgray!30}84.23 & \cellcolor{lightgray!30}86.41 & \cellcolor{lightgray!30}56.10 & \cellcolor{lightgray!30}69.42 & \cellcolor{lightgray!30}65.87 & \cellcolor{lightgray!30}71.90 & \cellcolor{lightgray!30}63.47 & $ \  \; \cellcolor{lightgray!30}\textbf{74.77}_{(\textcolor{blue}{\uparrow \textbf{1.60}})}$ \\
    \cmidrule{1-18}
    
    \multirow{2}{*}{CFA~\cite{guo2021beyond}}
	& R50 & 88.85 & 75.31 & 50.68 & 68.27 & 79.83 & 74.61 & 86.43 & 90.85 & 80.67 & 85.11 & 50.29 & 61.05 & 64.99 & 67.00 & 16.65 & $\ $ 69.37 \\
    & \cellcolor{lightgray!30}{\bf {\bf ARC-R50}} & \cellcolor{lightgray!30}88.96 & \cellcolor{lightgray!30}79.77 & \cellcolor{lightgray!30}52.08 & \cellcolor{lightgray!30}75.32 & \cellcolor{lightgray!30}79.46 & \cellcolor{lightgray!30}74.79 & \cellcolor{lightgray!30}86.98 & \cellcolor{lightgray!30}90.87 & \cellcolor{lightgray!30}80.90 & \cellcolor{lightgray!30}86.07 & \cellcolor{lightgray!30}58.16 & \cellcolor{lightgray!30}66.85 & \cellcolor{lightgray!30}66.15 & \cellcolor{lightgray!30}69.90 & \cellcolor{lightgray!30}46.74 & $ \  \; \cellcolor{lightgray!30}\textbf{73.53}_{(\textcolor{blue}{\uparrow \textbf{4.16}})}$ \\
    \cmidrule{1-18}
    
    Oriented
	& R50 & 89.48 & 82.59 & 54.42 & 72.58 & 79.01 & 82.43 & 88.26 & 90.90 & 86.90 & 84.34 & 60.79 & 67.08 & 74.28 & 69.77 & 54.27 & $\ $ 75.81 \\
    R-CNN~\cite{xie2021oriented} & \cellcolor{lightgray!30}{\bf ARC-R50} & \cellcolor{lightgray!30}89.40 & \cellcolor{lightgray!30}82.48 & \cellcolor{lightgray!30}55.33 & \cellcolor{lightgray!30}73.88 & \cellcolor{lightgray!30}79.37 & \cellcolor{lightgray!30}84.05 & \cellcolor{lightgray!30}88.06 & \cellcolor{lightgray!30}90.90 & \cellcolor{lightgray!30}86.44 & \cellcolor{lightgray!30}84.83 & \cellcolor{lightgray!30}63.63 & \cellcolor{lightgray!30}70.32 & \cellcolor{lightgray!30}74.29 & \cellcolor{lightgray!30}71.91 & \cellcolor{lightgray!30}65.43 & $\ $ \cellcolor{lightgray!30}\textbf{77.35}$_{(\textcolor{blue}{\uparrow \textbf{1.54}})}$ \\
    \bottomrule
\end{tabular}
}
\end{center}
\vskip -0.1in
\caption{{\bf Experiment results on the DOTA dataset among various detectors}. In the backbone column, R50 stands for ResNet-50~\cite{he2016deep}, and ARC-R50 is the backbone network that replaces the $3\times3$ convolution in the last three stage of ResNet-50 with the proposed ARC module. We conduct experiments on a variety of popular rotated object detection networks, including both single-stage methods (Rotated RetinaNet~\cite{lin2017focal}, R3Det~\cite{yang2021r3det}, S$^{2}$ANet~\cite{han2021align} ) and two-stage approaches (Rotated  FasterR-CNN~\cite{ren2015faster}, CFA~\cite{guo2021beyond}, Oriented R-CNN~\cite{xie2021oriented}). Experimental results validate the effectiveness and compatibility of the proposed method on various oriented detectors.}
\label{tab:various}
% \vspace{-0.1cm}
\end{table*}

\subsection{Adaptive rotated convolution module}
\label{sec:met_module}

In a regular convolution layer, the same convolution kernel is used for all input images. In contrast, the convolution kernel is adaptively rotated according to different input feature maps in the proposed adaptive rotated convolution module. Considering that object instances in an image usually face multiple directions, we introduce a conditional computation mechanism to handle objects of multiple orientations in the ARC module. The ARC module has $n$ kernels ($W_1, \cdots, W_n$), each with a shape of $[C_{\text out}, C_{\text in}, k, k]$. Given the input feature ${\boldsymbol x}$, the routing function $f$ predicts a set of rotation angles ${\boldsymbol \theta}$ and combination weights ${\boldsymbol \lambda}$:

\vskip -0.05 in
\begin{equation}
  {\boldsymbol \theta}, {\boldsymbol \lambda} = {\boldsymbol f}({\boldsymbol x}).
  \label{eq:important}
\end{equation}

\noindent
The $n$ kernels first rotate individually according to the predicted rotation angle  $ {\boldsymbol \theta} = [\theta_1, \theta_2, \cdots, \theta_n]$,

\vskip -0.05 in
\begin{equation}
  W^{\prime}_i = \text{Rotate}(W_i; \theta_i), i=1, 2, \cdots, n, 
  \label{eq:rotate}
\end{equation}

\noindent
where $\theta_i$ denotes the rotation angle for $W_i$, $W^{\prime}_i$ is the rotated kernel, and $\text{Rotate}({\cdot})$ is the procedure described in \cref{sec:met_rotate}.
A naive usage of these rotated kernels is convolving them with the input feature maps separately and adding the output feature maps together in an element-wise manner

\vskip -0.05 in
\begin{equation}
\setlength{\abovedisplayskip}{-0.5pt}
  \boldsymbol{y} = \lambda_1 (W^{\prime}_1 * {\boldsymbol x}) + \lambda_2 (W^{\prime}_2 * {\boldsymbol x}) + \cdots + \lambda_n (W
  ^{\prime}_n * {\boldsymbol x}),
  \label{eq:moe}
\end{equation}

\noindent
where $ {\boldsymbol \lambda} = [\lambda_1, \lambda_2, \cdots, \lambda_n]$ is the combination weights predicted by the routing function, $*$ is the convolution operation, and $\boldsymbol{y}$ is the combined output feature maps. Inspired by the conditional parameterization technique~\cite{yang2019condconv}, the above formula \cref{eq:moe} can be written as

\vskip -0.05 in
\begin{equation}
  \boldsymbol{y} = (\lambda_1 W^{\prime}_1 + \lambda_2 W^{\prime}_2 + \cdots + \lambda_n W^{\prime}_n) * {\boldsymbol x},
  \label{eq:condconv}
\end{equation}

\noindent
which means convolving the input feature separately and adding the output of these features (\cref{eq:moe}) is equivalent to performing one convolution operation with the combined convolution weight of these kernels (\cref{eq:condconv}). This strategy increases the representation ability of the network for capturing the features of multiple oriented objects yet remains highly efficient because the heavy convolution computation only occurs once in \cref{eq:condconv} compared to in \cref{eq:moe}.

\begin{table}
\begin{center}
  \resizebox{0.47\textwidth}{!}{
  \begin{tabular}{l | c | l l l}
    \toprule
    Method & Backbone & AP$_\text{50}$ & AP$_\text{75}$ & mAP \\
    \midrule
    Rotated
    & R50 & 84.20 & 58.50 & 52.70 \\
    RetinaNet~\cite{lin2017focal}
    & \cellcolor{lightgray!30 }{\bf ARC-R50} 
    & \cellcolor{lightgray!30}{\bf 85.10} 
    & \cellcolor{lightgray!30}{\bf 60.20}
    & \cellcolor{lightgray!30}{\bf 53.97}$_{(\textcolor{blue}{\uparrow \textbf{1.27}})}$ \\
    \midrule
    \multirow{2}{*}{S$^{2}$ANet~\cite{han2021align}}
    & R50 & 89.70 & 65.30 & 55.65 \\
    & \cellcolor{lightgray!30}{\bf ARC-R50} 
    & \cellcolor{lightgray!30}{\bf 90.00} 
    & \cellcolor{lightgray!30}{\bf 67.40}
    & \cellcolor{lightgray!30}{\bf 57.77}$_{(\textcolor{blue}{\uparrow \textbf{2.12}})}$ \\
    \midrule
    Oriented
    & R50 & 90.40 & 88.81 & 70.55 \\
    R-CNN~\cite{xie2021oriented}
    & \cellcolor{lightgray!30 }{\bf ARC-R50} 
    & \cellcolor{lightgray!30}{\bf 90.41} 
    & \cellcolor{lightgray!30}{\bf 89.02}
    & \cellcolor{lightgray!30}{\bf 72.39}$_{(\textcolor{blue}{\uparrow \textbf{1.84}})}$ \\
    \bottomrule
  \end{tabular}
  }
\end{center}
\vskip -0.1in
\caption{{\bf Experiment results on the HRSC2016 dataset.} The proposed kernel rotation mechanism is also effective on the HRSC2016 dataset on some popular rotated object detectors.}
\label{tab:exp_hrsc}
\vspace{-0.4cm}
\end{table}

\subsection{Implementation details}
\label{sec:met_imp}

Since the proposed ARC module can conveniently serve as a plug-and-play module for any backbone network with convolutional layer, we build the proposed backbone network ARC-ResNet based on the commonly used ResNet~\cite{he2016deep}. For the following experiments, we replace all the $3 \times 3$ convolutions in the last three stages and keep the $1 \times 1$ convolution the same as before, since $1 \times 1$ convolution is rotational invariant. The ablation studies on the performance of replacing different parts of the network are illustrated in \cref{sec:exp_abl}. In addition, we scale down the learning rate of the proposed backbone network during training. This adjustment on the rotated object detection tasks helps avoid drastic changes in the predicted rotation angles.

\section{Experiments}
\label{sec:exp}

In this section, we empirically evaluate our proposed backbone network equipped with adaptive rotated convolution on various detection networks. We begin by introducing the detailed experiment setup, which includes datasets and training configurations, in \cref{sec:exp_setting}. Then the main results of our method with various rotated object detectors on two commonly used datasets are presented in \cref{sec:exp_various}. The comparison results with competing approaches are demonstrated in \cref{sec:exp_sota}. Finally, the ablation studies in \cref{sec:exp_abl} and the visualization results in \cref{sec:exp_vis} further validate the effectiveness of the proposed method.

\begin{table*}[ht]
\begin{center}
\resizebox{1.0\textwidth}{!}{
\setlength{\tabcolsep}{1.1mm}
\begin{tabular}{l|c|c c c c c c c c c c c c c c c|l}
    \toprule
    Method & Backbone & PL & BD & BR & GTF & SV & LV & SH & TC & BC & ST & SBF & RA & HA & SP & HC & $\ $ mAP \\
    \midrule
    \multicolumn{18}{l}{{\bf \emph{Single-stage methods}}} \\
    \midrule
% 		RetinaNet-O$^\dag$ & R50 & 88.67 & 77.62 & 41.81 & 58.17 & 74.58 & 71.64 & 79.11 & 90.29 & 82.18 & 74.32 & 54.75 & 60.60 & 62.57 & 69.67 & 60.64 & 68.43 \\
		DRN~\cite{pan2020dynamic} & H104 & 88.91 & 80.22 & 43.52 & 63.35 & 73.48 & 70.69 & 84.94 & 90.14 & 83.85 & 84.11 & 50.12 & 58.41 & 67.62 & 68.60 & 52.50 & $\ $ 70.70 \\
		R3Det~\cite{yang2021r3det} & R101 & 88.76 & 83.09 & 50.91 & 67.27 & 76.23 & 80.39 & 86.72 & 90.78 & 84.68 & 83.24 & 61.98 & 61.35 & 66.91 & 70.63 & 53.94 & $\ $ 73.79 \\
		PIoU~\cite{chen2020piou} & DLA34 & 80.90 & 69.70 & 24.10 & 60.20 & 38.30 & 64.40 & 64.80 & 90.90 & 77.20 & 70.40 & 46.50 & 37.10 & 57.10 & 61.90 & 64.00 & $\ $ 60.50 \\
		RSDet~\cite{qian2021learning} & R101 & 89.80 & 82.90 & 48.60 & 65.20 & 69.50 & 70.10 & 70.20 & 90.50 & 85.60 & 83.40 & 62.50 & 63.90 & 65.60 & 67.20 & 68.00 & $\ $ 72.20 \\
		DAL~\cite{ming2021dynamic} & R50 & 88.68 & 76.55 & 45.08 & 66.80 & 67.00 & 76.76 & 79.74 & 90.84 & 79.54 & 78.45 & 57.71 & 62.27 & 69.05 & 73.14 & 60.11 & $\ $ 71.44 \\
        S$^{2}$ANet~\cite{han2021align} & R50 & 89.30 & 80.11 & 50.97 & 73.91 & 78.59 & 77.34 & 86.38 & 90.91 & 85.14 & 84.84 & 60.45 & 66.94 & 66.78 & 68.55 & 51.65 & $\ $ 74.13 \\
	    G-Rep~\cite{hou2022g} & R101 & 88.89 & 74.62 & 43.92 & 70.24  & 67.26 & 67.26 & 79.80 & 90.87 & 84.46 & 78.47 & 54.59 & 62.60 & 66.67 & 67.98 & 52.16 & $\ $ 70.59 \\
    \midrule
    
    \multicolumn{18}{l}{{\bf \emph{Two-stage methods}}} \\
    \midrule
		ICN~\cite{azimi2018towards} & R101 & 81.36 & 74.30 & 47.70 & 70.32 & 64.89 & 67.82 & 69.98 & 90.76 & 79.06 & 78.20 & 53.64 & 62.90 & 67.02 & 64.17 & 50.23 & $\ $ 68.16 \\
		CAD-Net~\cite{zhang2019cad} & R101 & 87.80 & 82.40 & 49.40 & 73.50 & 71.10 & 63.50 & 76.60 & 90.90 & 79.20 & 73.30 & 48.40 & 60.90 & 62.00 & 67.00 & 62.20 & $\ $ 69.90 \\
		RoI Trans~\cite{ding2019learning} & R101 & 88.64 & 78.52 & 43.44 & 75.92 & 68.81 & 73.68 & 83.59 & 90.74 & 77.27 & 81.46 & 58.39 & 53.54 & 62.83 & 58.93 & 47.67 & $\ $ 69.56 \\
		SCRDet~\cite{yang2019scrdet} & R101 & 89.98 & 80.65 & 52.09 & 68.36 & 68.36 & 60.32 & 72.41 & 90.85 & 87.94 & 86.86 & 65.02 & 66.68 & 66.25 & 68.24 & 65.21 & $\ $ 72.61 \\
		G. Vertex~\cite{xu2020gliding} & R101 & 89.64 & 85.00 & 52.26 & 77.34 & 73.01 & 73.14 & 86.82 & 90.74 & 79.02 & 86.81 & 59.55 & 70.91 & 72.94 & 70.86 & 57.32 & $\ $ 75.02 \\
		FAOD~\cite{li2019feature} & R101 & 90.21 & 79.58 & 45.49 & 76.41 & 73.18 & 68.27 & 79.56 & 90.83 & 83.40 & 84.68 & 53.40 & 65.42 & 74.17 & 69.69 & 64.86 & $\ $ 73.28 \\
		CenterMap~\cite{wang2020learning} & R50  & 88.88 & 81.24 & 53.15 & 60.65 & 78.62 & 66.55 & 78.10 & 88.83 & 77.80 & 83.61 & 49.36 & 66.19 & 72.10 & 72.36 & 58.70 & $\ $ 71.74 \\
		FR-Est~\cite{fu2020point} & R101 & 89.63 & 81.17 & 50.44 & 70.19 & 73.52 & 77.98 & 86.44 & 90.82 & 84.13 & 83.56 & 60.64 & 66.59 & 70.59 & 66.72 & 60.55 & $\ $ 74.20 \\
		Mask OBB~\cite{wang2019mask} & R50  & 89.61 & 85.09 & 51.85 & 72.90 & 75.28 & 73.23 & 85.57 & 90.37 & 82.08 & 85.05 & 55.73 & 68.39 & 71.61 & 69.87 & 66.33 & $\ $ 74.86 \\
		ReDet~\cite{han2021redet} & ReR50 & 88.79 & 82.64 & 53.97 & 74.00 & 78.13 & 84.06 & 88.04 & 90.89 & 87.78 & 85.75 & 61.76 & 60.39 & 75.96 & 68.07 & 63.59 & $\ $ 76.25 \\
		AOPG~\cite{cheng2022anchor} & R101  & 89.14 & 82.74 & 51.87 & 69.28 & 77.65 & 82.42 & 88.08 & 90.89 & 86.26 & 85.13 & 60.60 & 66.30  & 74.05 & 67.76 & 58.77 & $\ $ 75.39 \\
		SASM~\cite{hou2022shape} & R50  & 86.42 & 78.97 & 52.47 & 69.84 & 77.30 & 75.99 & 86.72 & 90.89 & 82.63 & 85.66 & 60.13 & 68.25 & 73.98 & 72.22 & 62.37 & $\ $ 74.92 \\
		
        \midrule
        \multirow{3.7}{*}{Oriented }
        & R50 & 89.48 & 82.59 & 54.42 & 72.58 & 79.01 & 82.43 & 88.26 & 90.90 & 86.90 & 84.34 & 60.79 & 67.08 & 74.28 & 69.77 & 54.27 & $\ $ 75.81 \\
        & \cellcolor{lightgray!30}{\bf ARC-R50} & \cellcolor{lightgray!30}89.40 & \cellcolor{lightgray!30}82.48 & \cellcolor{lightgray!30}55.33 & \cellcolor{lightgray!30}73.88 & \cellcolor{lightgray!30}79.37 & \cellcolor{lightgray!30}84.05 & \cellcolor{lightgray!30}88.06 & \cellcolor{lightgray!30}90.90 & \cellcolor{lightgray!30}86.44 & \cellcolor{lightgray!30}84.83 & \cellcolor{lightgray!30}63.63 & \cellcolor{lightgray!30}70.32 & \cellcolor{lightgray!30}74.29 & \cellcolor{lightgray!30}71.91 & \cellcolor{lightgray!30}65.43 & $\ $ \cellcolor{lightgray!30}\textbf{77.35}$_{(\textcolor{blue}{\uparrow \textbf{1.54}})}$ \\
        \cmidrule{2-18}
        \multirow{1}{*}{R-CNN~\cite{xie2021oriented}}
        & R101 & 89.51 & 84.50 & 54.67 & 73.10 & 78.77 & 82.87 & 88.08 & 90.90 & 86.97 & 85.38 & 63.06 & 67.35 & 75.91 & 68.73 & 51.91 & $\ $ 76.11 \\
        & \cellcolor{lightgray!30}{\bf ARC-R101} & \cellcolor{lightgray!30}89.39 & \cellcolor{lightgray!30}83.58 & \cellcolor{lightgray!30}57.51 & \cellcolor{lightgray!30}75.94 & \cellcolor{lightgray!30}78.75 & \cellcolor{lightgray!30}83.58 & \cellcolor{lightgray!30}88.08 & \cellcolor{lightgray!30}90.90 & \cellcolor{lightgray!30}85.93 & \cellcolor{lightgray!30}85.38 & \cellcolor{lightgray!30}64.03 & \cellcolor{lightgray!30}68.65 & \cellcolor{lightgray!30}75.59 & \cellcolor{lightgray!30}72.03 & \cellcolor{lightgray!30}65.68 & $\ $ \cellcolor{lightgray!30}\textbf{77.70}$_{(\textcolor{blue}{\uparrow \textbf{1.59}})}$ \\
    \bottomrule
\end{tabular}
}
\end{center}
\vskip -0.1in
\caption{{\bf Experimental results on the DOTA dataset compared with state-of-the-art methods.} In the backbone column, H104 denotes the 104-layer hourglass network~\cite{yang2017stacked}, DLA34 refers to the 34-layer deep layer aggregation network~\cite{zhou2019objects}, R50 and R101 stand for ResNet-50 and ResNet-101~\cite{he2016deep}, respectively, ReR50 is proposed in ReDet~\cite{han2021redet} with rotation-equivariant operations, ARC-R50 and ARC-R101 is the proposed backbone network which replaces the 3$\times$3 convolutions in ResNets with the proposed adaptive rotated convolution.}
\label{tab:sota}
\vspace{-0.3cm}
\end{table*}

\begin{table}
\begin{center}
  \resizebox{0.4\textwidth}{!}{
  \begin{tabular}{l|c|l}
    \toprule
    Method & Backbone & mAP \\
    \midrule
    R3Det~\cite{yang2021r3det} & R152 & 76.47 \\
    SASM~\cite{hou2022shape} & RX101 & 79.17 \\
    S$^{2}$ANet~\cite{han2021align} & R50 & 79.42 \\
    ReDet~\cite{han2021redet} & ReR50 & 80.10 \\
    R3Det-GWD~\cite{yang2021rethinking} & R152 & 80.19 \\
    R3Det-KLD~\cite{yang2021learning} & R152 & 80.63 \\
    AOPG~\cite{cheng2022anchor} & R50 & 80.66 \\
    KFIoU~\cite{yang2022kfiou} & Swin-T & 80.93 \\
    RVSA~\cite{wang2022advancing} & ViTAE-B & 81.24 \\
    \midrule
    \multirow{2}{*}{Oriented R-CNN~\cite{xie2021oriented}}
    & R50 &  80.62 \\
    & \cellcolor{lightgray!30 }{\bf ARC-R50} 
    & \cellcolor{lightgray!30}{\bf 81.77}$_{(\textcolor{blue}{\uparrow \textbf{1.15}})}$ \\
    \bottomrule
  \end{tabular}
  }
\end{center}
\vskip -0.1in
\caption{{\bf Experiment results on the DOTA dataset.} The results are obtained under {\bf \emph{multi-scale training and testing}} strategies.}
\label{tab:exp_dotams}
\vspace{-0.4cm}
\end{table}

\subsection{Experiment settings}
\label{sec:exp_setting}
{\bf Datasets.} We evaluate the proposed methods on two widely used oriented object detection benchmarks, \ie, DOTA-v1.0~\cite{xia2018dota} and HRSC2016~\cite{liu2016ship}.

DOTA~\cite{xia2018dota} is a large-scale rotated object detection dataset containing 2806 photographs and 188282 instances with oriented bounding box annotations. The following fifteen object classes are covered in this dataset: Plane (PL), Baseball diamond (BD), Bridge (BR), Ground track field (GTF), Small vehicle (SV), Large vehicle (LV), Ship (SH), Tennis court (TC), Basketball court (BC), Storage tank (ST), Soccer-ball field (SBF), Roundabout (RA), Harbor (HA), Swimming pool (SP), and Helicopter (HC). The image size of the DOTA dataset is extensive: from 800$\times$800 to 4000$\times$4000 pixels. We crop the raw images into 1024$\times$1024 patches with a stride of 824, which means the pixel overlap between two adjacent patches is 200. With regard to multi-scale training and testing, we first resize the raw pictures at three scales (0.5, 1.0, and 1.5) and crop them into 1024$\times$1024 patches with the stride of 524. Following the common practice, we use both the training set and the validation set for training, and the testing set for testing. The mean average precision and the average precision of each category are obtained by submitting the testing results to the official evaluation server of the DOTA dataset.

HRSC2016~\cite{liu2016ship} is another widely-used arbitrary-oriented object detection benchmark. It contains 1061 images with sizes ranging from 300$\times$300 to 1500$\times$900. Both the training set (436 images) and validation set (181 images) are used for training and the remaining for testing. For the evaluation metrics on the HRSC2016~\cite{liu2016ship}, we report the COCO~\cite{lin2014microsoft} style mean average precision (mAP) as well as the average precision under 0.5 and 0.75 threshold (AP$_\text{50}$ and AP$_\text{75}$). In the data pre-processing procedure, we do not change the aspect ratios of images.

{\bf Implementation Details.} The results on DOTA and HRSC2016 are obtained  using the MMRotate~\cite{zhou2022mmrotate} toolbox, except that Oriented R-CNN~\cite{xie2021oriented} is implemented with OBBDetection codebase. On the DOTA dataset, we train all the models for 12 epochs. On the HRSC2016 dataset, the Rotated RetinaNet~\cite{lin2017focal} is trained for 72 epochs, while S$^{2}$ANet~\cite{han2021align} and Oriented R-CNN~\cite{xie2021oriented} are trained for 36 epochs. The detailed configuration of various detectors is provided in the supplementary material. Although the detailed training configuration varies among different detectors and different datasets, we keep it the same between the experiments using our proposed backbone networks and baseline backbone networks for fair comparisons.

\subsection{Effectiveness on various architectures}
\label{sec:exp_various}

We compare the backbone network equipped with our proposed ARC module with counterparts that use a canonical ResNet-50~\cite{he2016deep}. The experiment results on DOTA~\cite{xia2018dota} dataset and HRSC2016~\cite{liu2016ship} dataset are shown in \cref{tab:various} and \cref{tab:exp_hrsc}, respectively. From the results, we find that our method significantly improves the generalization ability of various rotated object detection networks. On the most popular oriented object detection benchmark DOTA-v1.0, the proposed method achieves significant improvement on both single-stage and two-stage detectors. 

For single-stage detectors, our method could improve 3.03\% mAP for Rotated RetinaNet~\cite{lin2017focal}, 2.62\% mAP for R3Det~\cite{yang2021r3det}, and 1.36\% mAP for S$^{2}$ANet~\cite{han2021align}. For two-stage detectors, our method could also get more than 1.5\% mAP improvement (+1.60\% mAP for Rotated Faster R-CNN~\cite{ren2015faster}, +4.16\% mAP for CFA~\cite{guo2021beyond}
, and +1.54\% mAP for Oriented R-CNN~\cite{xie2021oriented}). On the HRSC2016~\cite{liu2016ship} dataset, the proposed method could also achieve remarkable improvement (1.27\% mAP improvement for  Rotated RetinaNet~\cite{lin2017focal}, 2.12\% for S$^{2}$ANet~\cite{han2021align}, and 1.84\% for Oriented R-CNN~\cite{xie2021oriented}). These experimental results verify that the proposed backbone networks are compatible with various detection network architectures and can effectively improve their performance on the oriented object detection tasks.

\begin{figure}[!t]
  \centering
  \includegraphics[width=1.0\linewidth]{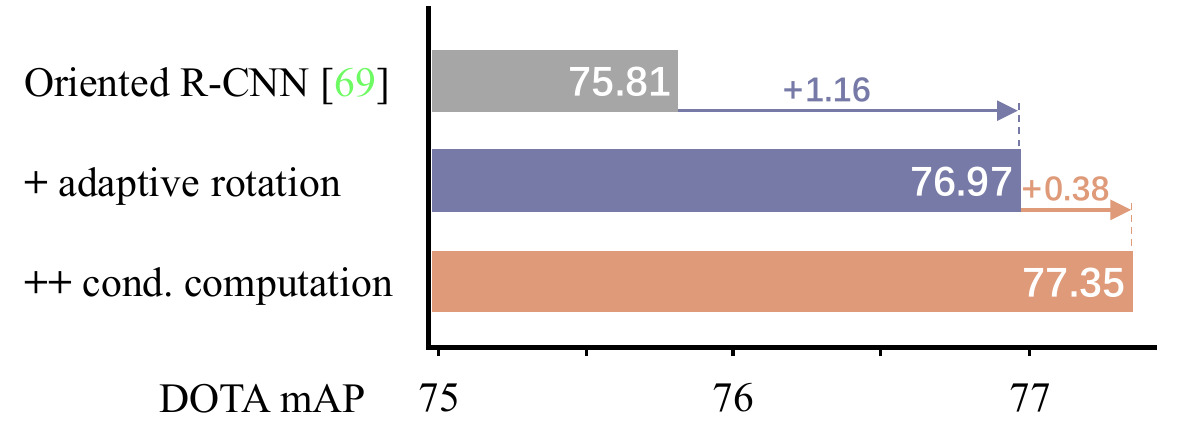}
  \caption{Ablation studies on the influence of adaptive kernel rotation and the effect of conditional computation. The experiments are conducted on Oriented R-CNN with DOTA dataset. }
  \label{fig:abl_rotate_experts}
  \vskip -0.1 in
\end{figure}

% The first observation is that under the same number of parameters, the employment of adaptive kernel rotation boosts the performance of rotated object detectors. The second observation is that the performance of the detector further increases with the adoption of conditional computation mechanism.

\subsection{Comparison with state-of-the-art methods.}
\label{sec:exp_sota}

We report full experimental results, including the average precision of each category and the mean average precision (mAP) on the DOTA dataset to make a fair comparison with the previous methods. We combine the proposed backbone network with one of the highly competitive method Oriented R-CNN~\cite{xie2021oriented}. The single-scale and the multi-scale training and testing results are shown in \cref{tab:sota} and \cref{tab:exp_dotams}, respectively. When we use the ARC-ResNet-50 backbone, the adaptive rotated convolution can improve the mAP of Oriented R-CNN by 1.54\% over the static convolution under the single scale training and testing strategy. As the depth of backbone network goes deeper to 101, the adaptive rotated convolution can still enhance the mAP by 1.59\%. Under the multi-scale training and multi-scale testing strategy, our method reaches 81.77\% mAP with ResNet-50 as the base model. This result is highly competitive and surpasses all other existing methods, even when compared with counterparts with vision transformer backbone~\cite{yang2022kfiou, wang2022advancing} or with advanced model pretraining mechanism~\cite{he2022masked, wang2022advancing}.

\subsection{Ablation studies}
\label{sec:exp_abl}

We conduct ablation studies to analyze how different design choices affect the rotated object detection performance. We first demonstrate that the adaptive kernel rotation mechanism significantly outperforms the static convolution approach. We further present the effectiveness of the conditional computation mechanism and the ablation studies on kernel number $n$. We then examine the effect of replacing different stages of the backbone network. Finally, the architecture design of our routing function is studied.

\begin{table}[t]
  \centering
  \resizebox{0.99\linewidth}{!}
  {
  \begin{tabular}{c|c|ccc|c}
    \toprule
    Backbone & $\; n \;$ & Params(M) & FLOPs(G) & FPS (img/s) & mAP \\  % Params(M) 
    \midrule
    % \multirow{5}{*}{Oriented }
    R50  & 1 & 41.14 & 211.43 & 29.9 & 75.81 \\  % 25.56
    R101 & 1 & 60.13 & 289.33 & 27.6 & 76.11 \\  % 44.55
    \cmidrule{1-6}
    % \multirow{2}{*}{R-CNN}  %~\cite{xie2021oriented}
    ARC-R50 & 1 & 41.18 & 211.85 & 29.6 & 76.97 \\  % params=41.14+0.0449400, flops=211.43 + 0.377494272 + 0.0417792, acts=303.07-232.26+259.52
    ARC-R50 & 2 & 52.25 & 211.89 & 29.2 & 77.17 \\  % params=41.14+11.111064, flops=211.43 + 0.377501184 + 0.0835584, acts=303.07-232.26+259.52
    ARC-R50 & 4 & 74.38 & 211.97 & 29.2 & 77.35 \\  % params=41.14+33.243312, flops=211.43 + 0.377515008 + 0.1671168, acts=303.07-232.26+
    ARC-R50 & 6 & 96.52 & 212.06 & 29.1 & 77.38 \\  % params=41.14+55.375560, flops=211.43 + 0.377528832 + 0.2506752, acts=303.07-232.26+
    % & ARC-R50 & 8 & 119.58 & 212.65 & 330.33 & todo \\  % params=41.14+, flops=211.43-86.578+87.794, acts=303.07-232.26+
    \bottomrule
  \end{tabular}
  }
  \vskip 0.15 cm
  \caption{Ablation studies on the kernel number $n$. The experiments are conducted on Oriented R-CNN with DOTA dataset.}
  \label{tab:ablation_number_kernel}
  \vskip -0.4 cm
\end{table}

\begin{figure*}[!t]
  \centering
  \includegraphics[width=1.00\linewidth]{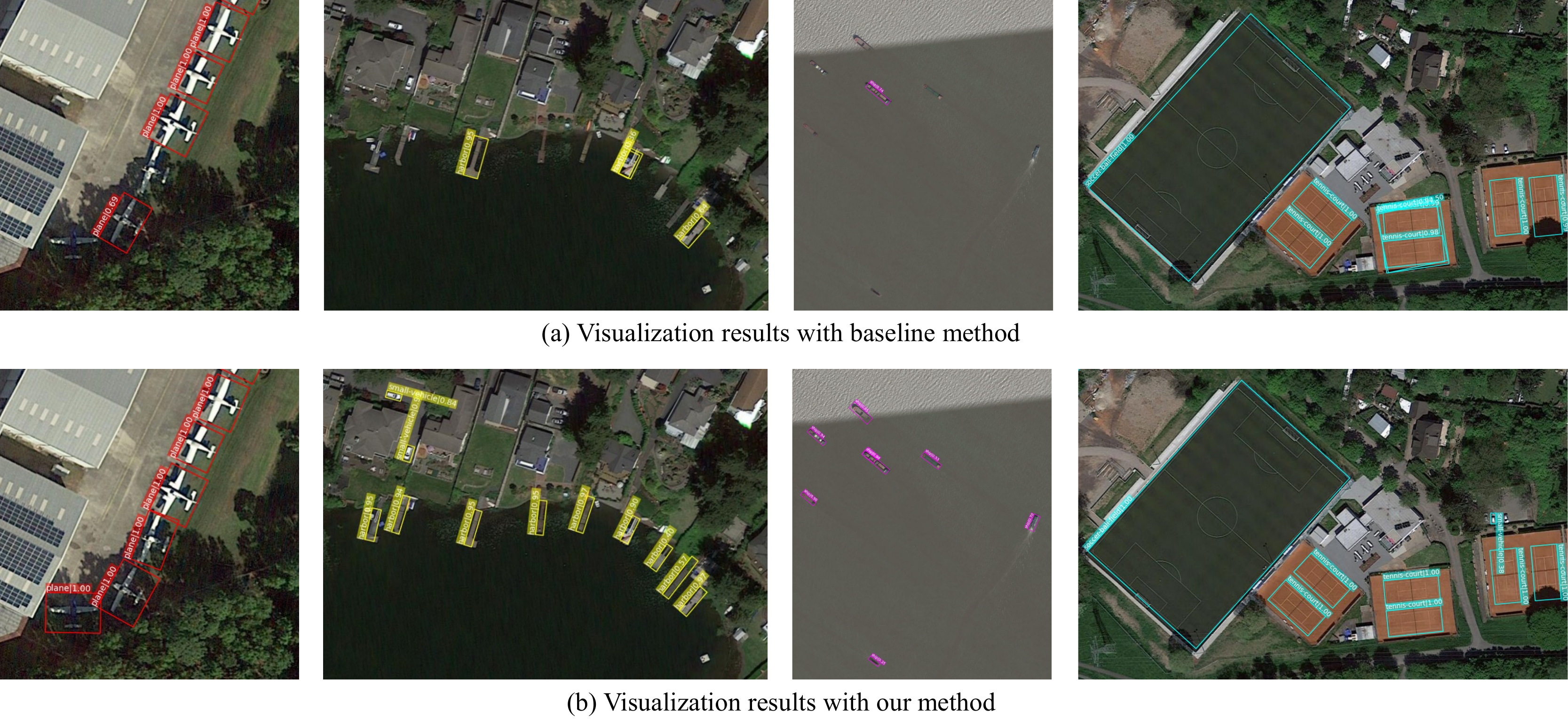}
  \caption{Visualization results of the oriented object detection on the test dataset of DOTA~\cite{xia2018dota}. The upper row displays the bounding boxes predicted by Oriented R-CNN~\cite{xie2021oriented} with ResNet-50 backbone as baseline. Right column shows the predictions of the Oriented R-CNN equipped with our proposed ARC module. \textbf{Zoom in for best view.}}
  \label{fig:vis_box}
  \vspace{-0.1cm}
\end{figure*}

{\bf Adaptive kernel rotation.} First, wecompare the performance between rotating the convolution kernels in a data-dependent manner and the static convolution on the DOTA dataset. The experiment results are shown in \cref{fig:abl_rotate_experts}. From the comparison between the first two rows in \cref{fig:abl_rotate_experts}, we find that by employing the adaptive kernel rotation, the performance of the object detector boosts (+1.16\% mAP on Oriented R-CNN~\cite{xie2021oriented}, which is highly competitive with SOTA methods). This experiment verifies the effectiveness of the proposed adaptive convolution kernel rotation approach for its adaptability  on capturing oriented objects.

{\bf Conditional computation.} 
The results in row 2 and row 3 of \cref{fig:abl_rotate_experts} show the effect of adopting more adaptive rotated kernels. By endowing the convolution kernel with more direction of rotation, the performance of the oriented object detector further increases. This is because the objects in an image usually have multiple orientations, and a single adaptive kernel is insufficient. Adopting more kernels with different orientation angles can endow the backbone feature extractor with more flexibility to produce high-quality features of these arbitrarily oriented objects.

{\bf Ablation on kernel number.}
\cref{tab:ablation_number_kernel} reports the information on the parameters, FLOPs, and inference speed (in FPS) and performance (in mAP) of different kernel numbers $n$.
The FLOPs are calculated with $1024 \times 1024$ image resolution.
The FPS is tested on an RTX 3090 with batch size 1, utilizing FP16 and torch.compile().
The results demonstrate that, as we progressively increase the number of kernels, the mAP exhibits a consistent upward trend. Meanwhile, FLOPs and FPS are essentially unchanged.
This phenomenon demonstrate that the proposed method achieves remarkable improvement in mAP while maintaining a high level of efficiency, with only a marginal increase of 0.002\% in FLOPs and less than 2.7\% drop in FPS compared to the baseline model.
Note that the number of parameters is no longer the bottleneck for evaluating model efficiency.

\begin{table}[ht]
\begin{center}
  \begin{tabular}{c c c c|l}
    \toprule
    Stage1 & Stage2 & Stage3 & Stage4 & mAP \\
    \midrule
    - &   -    &   -    & - & 75.81 \\
    - &   -    &   -    & \cmark & 77.17$_{(\textcolor{blue}{\uparrow \textbf{1.36}})}$ \\
    - &   -    & \cmark & \cmark & 77.29$_{(\textcolor{blue}{\uparrow \textbf{1.48}})}$\\
    - & \cmark & \cmark & \cmark & {\bf 77.35}$_{(\textcolor{blue}{\uparrow \textbf{1.54}})}$ \\
    \bottomrule
  \end{tabular}
\end{center}
\vskip -0.1in
\caption{Ablation studies of the replacement strategy on the DOTA dataset. The experiments are conduct on Oriented R-CNN with ARC-ResNet-50 backbone network, which have 3, 4, 6, 3 blocks on each stage, respectively. The symbol `\cmark' means we replace all the convolution layers in this stage, while the symbol `-' means we do not replace any convolution blocks in this stage. }
\label{tab:replace}
\vskip -0.2in
\end{table}

{\bf Replacement strategy.} Since the feature pyramid network (FPN)~\cite{lin2017feature} is attached to the last three stages of the backbone network, we do not replace the convolution layer in the first stage and ablate the replacement strategy in the last three stages. The ablation experiments are conducted on Oriented R-CNN~\cite{xie2021oriented} with a 50-depth backbone network on the DOTA dataset. Initially, we replace all the 3$\times$3 convolution with the proposed ARC module in the last stage of the backbone network, and the mAP metric gets a 1.36\% improvement over the baseline model. We further replace the convolution layer of more stages at the backbone network, and the performance on the oriented object detection benchmark improves steadily. As a result, we choose to replace all the last three stages in the backbone network.

{\bf The structure of the routing function.} We ablate two designs in the routing function. The first design is spatial information encoding, which adds a depthwise convolution module before the average pooling layer (see \cref{fig:module}{\color{red} (c)}). The second design uses the combination weights prediction branch to adaptively combine the weight of each kernel (the branch producing $\boldsymbol{\lambda}$ in \cref{fig:module}{\color{red} (c)}) rather than simply taking the average value of them. We conduct the experiments with Oriented R-CNN~\cite{xie2021oriented} on the DOTA dataset~\cite{xia2018dota}, and the corresponding experiment results are shown in \cref{tab:router}. When we add the spatial encoding modules, the performance can increase from 76.41\% to 76.80\%. This is because the additional convolution layer helps the routing function to capture the spatial orientation information from the feature maps. Meanwhile, the introduction of the adaptive combination could also get a 0.47\% reward in mAP, which shows the advantage of adopting the adaptiveness among different rotated kernels. When we use both of the designs, the oriented object detector achieves the highest performance. As a result, we choose to adopt both of the two designs into our proposed routing function.

\begin{table}[t]
\begin{center}
  \resizebox{0.45\textwidth}{!}{
  \begin{tabular}{c|c c c c}
    \toprule
    adaptive combination & \xmark & \xmark & \cmark & \cmark \\
    spatial info. encoding     & \xmark & \cmark & \xmark & \cmark \\
    \midrule
    mAP & 76.41 & 76.80 & 76.98 & \textbf{77.35} \\
    \bottomrule
  \end{tabular}
  }
\end{center}
\vspace{-0.2cm}
\caption{Ablation studies on the structure of the routing function. \emph{Adaptive combination} means using the combination weights prediction branch to adaptively combine the weight of each kernel, and \emph{spatial encoding} refers to adding a DWConv-LN-ReLU module before the average pooling layer.}
\label{tab:router}
\vspace{-0.4cm}
\end{table}

\subsection{Visualization}
\label{sec:exp_vis}

To give a deep understanding of our method, we visualize the predicted oriented bounding boxes and the corresponding scores. The experiment is conducted with Oriented R-CNN~\cite{xie2021oriented} detector on the test set of DOTA. By comparing the results between the detector with our proposed backbone and that with the baseline backbone in \cref{fig:vis_box}, the proposed method shows its superiority. To be specific, thanks to its adaptiveness in handling arbitrarily oriented object instances, our method has a superior locating and identification ability on both small (\eg ship in the third column), and median (\eg plane and harbor in the first two columns), as well large (\eg tennis court and soccer ball field in the last column) oriented object instances.

\section{Conclusion}
\label{sec:conclusion}

This paper proposed an adaptive rotated convolution module for rotated object detection. In the proposed approach, the convolution kernels rotate adaptively according to different object orientations in the images. An efficient conditional computation approach is further introduced to endow the network with more flexibility to capture the orientation information of multiple oriented objects within an image. The proposed module can be plugged into any backbone networks with convolution layer. Experiment results verify that, equipped with the proposed module in the backbone network, the performance of various oriented object detectors improves significantly on commonly used rotated object detection benchmarks while remaining efficient.

\vskip 0.2 cm
\noindent\textbf{Acknowledgement.} This work is supported in part by the National Key R\&D Program of China under Grant 2021ZD0140407, the National Natural Science Foundation of China under Grants 62022048 and 62276150, THU-Bosch JCML and Guoqiang Institute of Tsinghua University. We also appreciate the generous donation of computing resources by High-Flyer AI.

{\small
\bibliographystyle{ieee_fullname}
\bibliography{references}
}

\end{document}